\documentclass[conference]{IEEEtran}
\usepackage{float}
\usepackage{placeins}
\usepackage{subfig}

\ifCLASSINFOpdf
  \usepackage[pdftex]{graphicx}
\else
\fi
\usepackage{amsmath}
\usepackage[utf8]{inputenc}
\usepackage[linesnumbered,ruled]{algorithm2e}
\usepackage{fixltx2e}
\usepackage{dblfloatfix}

\usepackage{booktabs}
\usepackage{bm}

\begin{document}
%
\title{An Ensemble Generation Method\\ Based on Instance Hardness}

\author{\IEEEauthorblockN{
    Felipe N. Walmsley,
    George D. C. Cavalcanti\\
    and
    Dayvid V. R. Oliveira
    }
    \IEEEauthorblockA{
        Centro de Inform\'atica - Universidade Federal de Pernambuco\\
        Email: \{fnw, gdcc, dvro\}@cin.ufpe.br
    }
    \and
    \IEEEauthorblockN{
        Rafael M. O. Cruz
        and
        Robert Sabourin
    }
    \IEEEauthorblockA{
        École de Technologie Sup\'erieure - Universit\'e du Qu\'ebec\\
        Email: \{cruz, robert.sabourin\}@livia.etsmtl.ca
    }
}


%


\maketitle

\begin{abstract}
In Machine Learning, ensemble methods have been receiving a great deal of attention. Techniques such as Bagging and Boosting have been successfully applied to a variety of problems. Nevertheless, such techniques are still susceptible to the effects of noise and outliers in the training data. We propose a new method for the generation of pools of classifiers based on Bagging, in which the probability of an instance being selected during the resampling process is inversely proportional to its instance hardness, which can be understood as the likelihood of an instance being misclassified, regardless of the choice of classifier. The goal of the proposed method is to remove noisy data without sacrificing the hard instances which are likely to be found on class boundaries. We evaluate the performance of the method in nineteen public data sets, and compare it to the performance of the Bagging and Random Subspace algorithms. Our experiments show that in high noise scenarios the accuracy of our method is significantly better than that of Bagging.
\end{abstract}


%
\IEEEpeerreviewmaketitle

\section{Introduction}\label{sec:introduction}

\subsection{Motivation}

Ensemble methods \cite{zhou2012ensemble} \cite{cruz2018dynamic} \cite{Britto2014} are techniques that combine multiple predictors trained independently, using a combination of the outputs of each predictor as the final output. This is in contrast to traditional Machine Learning methods, which train a single classifier on the whole of the training set. The rationale behind this shift in paradigm is that by using an ensemble of classifiers one can expect to obtain a pool of predictors with complementary competences - that is, the predictors make correct and incorrect predictions on different patterns, complementing one another. The pools generated can thus obtain performance gains over strategies that employ a single classifier, given that finding the single optimal model for a problem may be exceedingly difficult.

One ensemble learning method that has enjoyed widespread adoption is the Bootstrap Aggregating Algorithm (Bagging) \cite{breiman1996bagging}. The algorithm relies on creating an ensemble of $N$ classifiers trained on $N$ training sets created from the original training set. These training sets are generated by sampling uniformly and with replacement from the original training set.

The use of Bagging is interesting when the data sets available are small, noisy, or both \cite{khoshgoftaar2011comparing}. In general terms, it can be expected that the classifiers produced by the Bagging algorithm will have complementary competences, making the decisions of the system better than those of a single classifier trained on the whole training set \cite{breiman1996bagging}.

Nevertheless, while ensemble learning may offer performance gains, they cannot completely avoid two common problems in machine learning: noisy data and outliers.  In this work, we call ``outliers" those examples that are considerably different from most members of its class, while the word ``noise'' is used to refer to sources of noise that cannot be removed by calibration. In this work, we focus on label noise, as opposed to noise that acts on the features of instances. By label noise we mean any process that changes the label of an instance, as presented to a learning algorithm, from its true value.

The main issue related to the presence of noise, outliers, or both, is that the training process of a classifier can become unstable or prone to overfitting \cite{long2010random}, regardless of the use of ensemble techniques. This is particularly concerning in the case of algorithms that place greater weight on misclassified instances during the training process, such as AdaBoost \cite{Schapire1999}, since it is possible that the model will be strongly adjusted so as to correctly classify instances that do not represent the underlying distribution of the data. In these scenarios, one would expect to observe a significant decrease in the generalization accuracy of the model. A rather comprehensive treatment of label noise and its negative impact on classification accuracy can be found in \cite{Frenay2014}.

Nevertheless, there are techniques that aim to remove noisy instances from a data set, as a means to alleviate the previously mentioned problems. For a review of instance selection techniques, including those focused on removing noisy instances, see \cite{Olvera-Lopez2010} \cite{garcia2015data}. Still, noise removal techniques might cause undesired side effects on the training set, such as the removal of examples that are not noise or a dramatic removal of examples in the boundaries between classes.

Parallel to the concept of noise we have the concept of instance hardness \cite{smith2014instance}, or the difficulty in classifying an instance. The hardness of an instance can be understood as the likelihood that it will be incorrectly classified, regardless of the choice of learning algorithm. The difficulty in classifying an instance may be used as proxy to the probability that such instance is noisy or an outlier. In this manner, instance hardness measures may be used as a foundation for techniques which aim to selectively remove instances from the training set.

\subsection{Objective}

Armed with the concepts of outliers, noise and instance hardness, it is natural to question whether it would be possible to use some measure of instance hardness to remove the troublesome instances mentioned previously  from a training set. One would expect that once those instances are removed, the training process would become more stable, leading to better generalization accuracy. In \cite{smith2014instance}, the authors propose a filtering scheme based on instance hardness, in which the patterns with a hardness value above a certain threshold are removed before training. In order to calculate the hardness of a pattern, several classifiers are trained, and the confidence of each classifier in the classification of the pattern is measured. The less confident the classifiers are, the harder the instance.

This work proposes an ensemble generation method based on Bagging that seeks to remove outliers and noisy instances from the training set, while still preserving instances that are close to class borders.
The core idea of this method is to modify the process by which the bootstrapped training sets are created in the Bagging algorithm. Instead of picking examples with uniform probability, the probability of an instance being picked is now defined to be inversely proportional to its hardness. This probabilistic approach stands in contrast to hard filtering methods, and aims to reduce the frequency with which noisy instances are chosen, while still allowing for the selection of hard instances which may lie on the boundaries between classes, therefore preserving class boundary information.

\subsection{Methodology}

In order to assess the effectiveness of the proposed method, we compare it with Bagging and Random Subspace \cite{Ho1998}. We add artificial noise to the class labels of a data set by randomly shifting the class labels of the instances, and then train classifiers on this noisy data. The accuracy of each method is measured in 19 public datasets from the UCI \cite{Lichman2013} and KEEL \cite{alcala2011keel} repositories, and a statistical analysis of the results is performed, in order to evaluate if the proposed method achieves superior generalization accuracy.

Furthermore, we conduct an analysis of the frequency with which our proposed method adds noisy examples to the bootstrapped training sets. This is done in order to ascertain whether our method is truly capable of avoiding the selection of noisy instances, by distinguishing noisy examples from non-noisy ones.

\section{Proposed Method}\label{sec:proposed}

The method proposed in this work is based on Bagging \cite{breiman1996bagging}. Our motivation in proposing changes to Bagging is that, while it may offer accuracy gains, its sampling process is still subject to the effects of noisy data or outliers, given that all instances are equally likely to be selected.  Therefore, it is possible that some of the bootstrapped training sets might have a high proportion of noise, outliers or both. When presented with such training data, some classifiers may either overfit the noisy data or to fail to learn at all, thereby reducing the generalization accuracy of the final system.


To alleviate this issue, we propose a modification to the sampling process used in the Bagging algorithm. Our motivations in proposing this new method are twofold:
\begin{enumerate}
\item{We would like to avoid adding noisy examples to the bootstrapped sets too frequently.}
\item{We also would like to avoid completely removing ``hard'' instances, as they might be instances on the boundary of classes. }
\end{enumerate}

In order to deal with noisy examples, we must first devise a means to identify them. To that end, we estimate the probability of an instance being noise by applying instance hardness measures. More specifically, we use the k-Disagreeing Neighbors measure, introduced in \cite{smith2014instance} as a measure of the hardness of the instance. The k-Disagreeing Neighbors measure was chosen as the experiments in \cite{smith2014instance} indicate that it is correlated with the frequency with which an instance is misclassified.

The k-Disagreeing Neighbors (kDN) measure is defined as the fraction of the $k$ nearest neighbors of a sample that do not share its class label. Formally, the kDN hardness $kDN(x)$ of an instance $x$ , whose $k$ nearest neighbors are denoted by $kNN(x)$, is defined as:

\begin{equation}\label{eq:kdn}
kDN(x) = \frac{\left\vert x' \mid x' \in kNN(x) \land label(x') \neq label(x) \right\vert}{k}
\end{equation}
where $label(x)$ is the class label of example $x$.

From the definition, $kDN(x)$ takes on values in the interval $\lbrack 0,1\rbrack$, evenly spaced by $\frac{1}{k}$.

We expect instances near the mean of a class to have a low instance hardness values, as they are mostly surrounded by instances of the same class. On the other hand, instances near the border of classes will have neighbors which belong to other classes, making their instance hardness larger. Nevertheless, we want to preserve this latter type of instance, which makes a hard filtering scheme unsuitable. Thus, we adopt a probabilistic approach, in which examples are chosen during the bootstrapping process with a probability that is larger the smaller its hardness is. This scheme results in a low selection probability for noisy instances, while still allowing for the hard instances near the border of classes to be picked with non-zero probability.

Algorithm \ref{alg:method} presents the pseudocode of the proposed technique, which we dub ``Bagging-IH''.
Lines 3 to 5 of Algorithm \ref{alg:method} show the process of calculating the hardness of the instances, according to Equation \ref{eq:kdn}.

\begin{algorithm}[]
\caption{Bagging-IH: The pool generation algorithm}
\label{alg:method}
\SetKwInOut{Input}{Input}\SetKwInOut{Output}{Output}

\Input{The training set $T$\\
The pool size $m$\\
The bootstrapped set size $n_{b}$\\
The base predictor $C$\\
The value of k for the kDN measure\\
A boolean $p_{type}$
}
\Output{The trained pool $P$}
\Begin{
Initialize the pool $P$ as the empty set

\ForEach{$x \in T$}
{
  $h(x)  = kDN(x)$
}

\ForEach{$x \in T$}
{
     $p(x)$ = $normalize(x, h(x), T)$ \tcc*{The selection probability of the instance}
}

\For{i from 1 to $m$}
{
Initialize the training set $T_{i}$ as the empty set

	\For{j from 1 to $n_{b}$}
    {
    	Add an instance $x_{j} \in T$ to $T_{i}$, sampled with replacement according to $p(x_{i})$
    }

Train the classifier $C_{i}$ (an instance of $C$ ) using $T_{i}$ \\
Add $C_{i}$ to the pool $P$
}

}
\end{algorithm}

Once the hardness of each instance is calculated, the procedure for calculating probabilities returns a selection likelihood which is inversely proportional to the instance hardness. Let $n$ be the number of examples in the training set $T$. We define the function $f(x_{i})$ of an instance $x_{i} \in T$:

\begin{equation}\label{eq:prior}
f(x_{i}) = \frac{1}{n} + (1 - kDN(x_{i}))
\end{equation}

The first term in Equation \ref{eq:prior} attributes a uniform selection likelihood to all instances of the training set. This ensures that even instances with a kDN value of 1 are given a chance to be selected, as an attempt to avoid the issue of instances that are altogether discarded. The rationale behind this decision is to preserve possible hard instances in the boundary of classes. Meanwhile, the second term makes it so that harder instances are less likely to be selected.

We then normalize (line 7 of Algorithm \ref{alg:method}) the value of $f(x_{i})$ by the sum over all $x_{i} \in T$ to obtain the probability of the instance being selected, $p(x_{i})$, according to Equation \ref{eq:plinear}.

\begin{equation}\label{eq:plinear}
p(x_{i}) = \frac{f(x_{i})}{\sum\limits_{i=1}^n f(x_{i})}
\end{equation}

The normalization is used to ensure that $p$ is a proper probability distribution. This is necessary for the proper functioning of computational implementations of the procedure of choosing with replacement.

Once the selection probabilities have been calculated, we proceed to generate the bootstrapped training sets as in the original Bagging algorithm. Drawing with replacement from a probability distribution given by Equation \ref{eq:plinear}, $m$ bootstrapped training sets are generated (lines 11-13 of Algorithm \ref{alg:method}), and $m$ instances of a base predictor are trained (lines 14 and 15 of Algorithm \ref{alg:method}).
By making the selection probability inversely proportional to the instance hardness, we aim to pick “hard” examples with a smaller probability than that of “easy” examples during the bootstrapping process.

It is important to note that while the original Bagging algorithm was concerned with predictors in general, i.e. both classifiers and regressors, the work here presented is focused on classifiers. This means that we are dealing with pools of classifiers, and that the output of the ensemble at test time will be a class label. It should be noted that this restriction is due to our choice of hardness measure. It should be possible to adapt the method for regression, by using a measure of instance hardness that does not depend on the class of the instance as the ones in \cite{smith2014instance} do.

At test time, the predicted class label for an instance is calculated according to Algorithm \ref{alg:test}. Each classifier in the trained pool outputs a class prediction for the test instance $x$. The predictions are then weighted under some voting scheme (e.g. Majority Vote) to give the final output of the pool, $y_{pred}(x)$. The value $y_{pred}(x)$ is the one considered when evaluating the performance of the pool.

\begin{algorithm}[]
\caption{Prediction on a test instance}
\label{alg:test}
\SetKwInOut{Input}{Input}\SetKwInOut{Output}{Output}

\Input{The trained pool $P$\\
The size of the pool $m$\\
A test instance $x$\\
A voting scheme $V$ for the predictions of the pool
}
\Output{The predicted class label of $x$, $y_{pred}(x)$}
\Begin{

Initialize the set of pool predictions $Y$ to the empty set

\For{i from 1 to $m$}
{
    Calculate the class label $y_{i}(x)$ of $x$ predicted by classifier $C_{i}$ in the pool.\\
    Add $y_{i}(x)$ to $Y$
}

Calculate $y_{pred}(x)$ as the result of $V(Y)$

}
\end{algorithm}

\section{Experiments and Discussion}\label{sec:experiments}


In this section, we compare the performance of the proposed method against that of commonly used pool generation methods. We also evaluate whether the proposed is actually capable of discerning noisy instances from non-noisy instances.

\subsection{Experimental design}

In order to assess the effectiveness of pool generation methods, we compare their accuracy on several public data sets, which are described shortly. The proposed method was compared with the Bagging and Random Subspace algorithms. In order to obtain a baseline, the performance of a single monolithic classifier was also evaluated. This classifier uses the same algorithm as the base classifier used to compose the pools.

We adopted a 5-fold cross-validation approach to partition the data sets, with 4 of the folds in each partition being used for training and the last for testing. The mean accuracy over the folds for each pool generation algorithm was measured. The cross-validation procedure was repeated 20 times for each noise scenario analyzed, and the average and standard deviation of the mean accuracy were measured.

\subsubsection{Experimental parameters}

For all the pool generation methods, we used a pool size $m = 50$, the Perceptron as the base classifier $C$, a bootstrapped set size $n_{b} = \left \vert T \right \vert$ (where $T$ is the training set, and $\left \vert T \right \vert$ is the number of examples in $T$) and the Majority Vote Rule as the voting scheme. The Majority Vote Rule was chosen in order to avoid introducing confounding factors into the performance analysis. By using a simple combination rule that is applicable to most ensemble methods, we can focus on evaluating the differences in performance caused by the choice of the pool generation procedure. The Perceptron algorithm was chosen for being an unstable learner, and also for generating an easy to interpret decision rule. The Perceptrons were trained in a One-vs-All scheme for the multi-class problems. The pool size was chosen based on the results reported by Roy et al. in \cite{roy2016neurocomp}, \cite{royicpr}, which indicate that it is rarely necessary to use a pool size of more than 50 to achieve the best results for a given ensemble algorithm. For the experiments involving the kDN measure, a value of $k = 5$ was chosen.

As a baseline for comparison, in our experiments we also train a single monolithic classifier on the entire training set, using the scheme described above. We dub this classifier ``Perceptron OvA''.

There is one extra parameter for the Random Subspace algorithm, which is the maximum size of the reduced feature set. In our work, we set the maximum size to be 50\% of the original feature space dimension. In \cite{Ho1998}, the author found that using about half the features would result in the best accuracy for the case under study. Nevertheless, since this value is likely to be dependent on the data set, and we do not perform this sort of hyperparameter optimization for any other methods, we opted to use the same value for the percentage of features kept for all data sets.

\subsubsection{Data sets}

The public data sets used in our experiments are described in Table \ref{tbl:datasets}. All data sets were obtained from the UC Irvine Machine Learning Repository \cite{Lichman2013}, except for the Glass (\#4) and Satimage (\#12) data sets, obtained from the KEEL-data set repository \cite{alcala2011keel}.

\begin{table}
  \caption{The data sets used in the experiments}
  \label{tbl:datasets}
\resizebox{\columnwidth}{!}{
\begin{tabular}{lllll}
\hline
\textbf{} & \textbf{Data set}             & \textbf{\# of examples} & \textbf{Dimensions} & \textbf{\# of classes} \\ \midrule
1         & Blood Transfusion             & 747                     & 4                   & 2                     \\
2         & CTG                           & 2126                    & 21                  & 3                      \\
3         & E. Coli                       & 336                     & 7                   & 8                      \\
4         & Glass                         & 214                     & 9                   & 6                      \\
5         & Haberman's Survival           & 306                     & 3                   & 2                      \\
6         & Indian Liver Patient Database & 579                     & 10                  & 2                      \\
7         & Ionosphere                    & 351                     & 34                  & 2                      \\
8         & Liver                         & 345                     & 6                   & 2                      \\
9         & New Thyroid                   & 215                     & 5                   & 3                      \\
10        & Page Blocks                   & 5473                    & 10                  & 5                      \\
11        & Pima                          & 768                     & 8                   & 2                      \\
12        & Satimage                      & 6435                    & 36                  & 6                      \\
13        & Segment                       & 2310                    & 19                  & 7                      \\
14        & Shuttle                       & 58000                   & 9                   & 7                      \\
15        & Vehicle                       & 846                     & 18                  & 4                      \\
16        & Vertebral Column              & 310                     & 6                   & 2                      \\
17        & Vowel                         & 990                     & 13                  & 11                     \\
18        & WDBC                          & 569                     & 30                  & 2                      \\
19        & Yeast                         & 1484                    & 8                   & 10                     \\ \hline
\end{tabular}
}

\end{table}

Since both the kNN algorithm and the Perceptron algorithm can be affected by the presence of features which have very different scales, we perform feature-wise scaling on all data sets, in order to have all features lie on the $[0,1]$ interval.

\subsubsection{Assessing the effect of noise}
One of the main motivations of this work is to investigate the effects of noise on the accuracy of pools of classifiers. The method we have proposed is heavily focused on dealing with the effects of noisy instances in the training data. Therefore, it is paramount that we conduct our experiments in a manner that allows us to systematically evaluate the effects of noise on the data.

In order to control the noise on the data sets and to be able to observe the behavior of the classifiers and pools under different noise conditions, we adopted the following procedure for adding noise to the data sets: For each instance $x \in T$, where $T$ is the training set, its class label has a probability $pr_{change}$ of being changed to one of the other classes present in the data set. Another way to put this is that the expected fraction of elements that will have its label changed is $pr_{change}$. In this work, we evaluated all algorithms for values of $pr_{change}$ in the set $\{0, 0.1, 0.2, 0.3, 0.4\}$. The case $pr_{change} = 0$ is the scenario in which no noise is added.

\subsubsection{Methods for statistical analysis}
In order to analyze the statistical significance of our results we use the Wilcoxon signed-rank test. Our null hypothesis is that, for each pairwise comparison, the classifiers are equivalent. More precisely, our null hypothesis is that the distribution of the difference of their accuracies is symmetric about zero. We chose a $p$-value of 0.05, therefore rejecting the null hypothesis if $p < 0.05$.

\subsection{Results}

In this section, we present the results of our experiments. The analyses in this section are grouped by noise scenario, as this makes it easier to understand how the proposed algorithm behaves under different circumstances. Following each set of results, we present our comments on them.
\subsubsection{Noise Free Scenario}

In order to obtain a baseline of results, our first evaluation is concerned with the performance of the classifiers under a noise-free scenario. Table \ref{tbl:acc0} shows the accuracy of each algorithm under this scenario. The last line of Table \ref{tbl:acc0} also shows the results of the two-tailed Wilcoxon signed-rank test.

\begin{table}
  \caption{Mean and standard deviation of the mean accuracy over the folds for each classifier, at a noise level of 0\%. The best values (highest mean) are shown in bold. The last line of the table shows the p-value for each pairwise Wilcoxon test.}
  \label{tbl:acc0}
\resizebox{\columnwidth}{!}{%
\begin{tabular}{lllll}
\toprule
              Dataset &                       \textbf{\textbf{Bagging}-IH} &                                \textbf{Bagging} &        \textbf{Rand. Subsp.} &                 \textbf{Perceptron OvA} \\
\midrule
 blood transfusion &  $ \bm{ 78.51 \pm 0.40 } $ &                  $ 78.44 \pm 0.43 $ &  $ 70.77 \pm 3.28 $ &  $ 67.36 \pm 4.81 $ \\
               ctg &                  $ 88.05 \pm 0.22 $ &  $ \bm{ 88.50 \pm 0.31 } $ &  $ 85.74 \pm 1.16 $ &  $ 85.21 \pm 1.22 $ \\
             ecoli &  $ \bm{ 77.88 \pm 0.97 } $ &                  $ 76.86 \pm 0.74 $ &  $ 60.30 \pm 4.24 $ &  $ 61.42 \pm 4.85 $ \\
             glass &  $ \bm{ 57.63 \pm 2.07 } $ &                  $ 57.42 \pm 2.86 $ &  $ 45.45 \pm 3.00 $ &  $ 46.12 \pm 3.48 $ \\
          haberman &  $ \bm{ 74.39 \pm 0.73 } $ &                  $ 73.84 \pm 0.62 $ &  $ 67.11 \pm 6.57 $ &  $ 68.24 \pm 5.24 $ \\
              ilpd &  $ \bm{ 70.75 \pm 0.80 } $ &                  $ 70.73 \pm 1.12 $ &  $ 63.79 \pm 5.47 $ &  $ 64.37 \pm 4.70 $ \\
        ionosphere &                  $ 84.79 \pm 0.74 $ &  $ \bm{ 86.87 \pm 0.89 } $ &  $ 76.73 \pm 7.15 $ &  $ 73.87 \pm 7.79 $ \\
             liver &                  $ 65.46 \pm 1.72 $ &  $ \bm{ 66.25 \pm 1.67 } $ &  $ 58.20 \pm 1.17 $ &  $ 59.70 \pm 2.37 $ \\
       new\_thyroid &                  $ 90.07 \pm 0.54 $ &  $ \bm{ 92.56 \pm 0.91 } $ &  $ 86.56 \pm 0.95 $ &  $ 89.70 \pm 1.49 $ \\
       page\_blocks &                  $ 94.11 \pm 0.15 $ &  $ \bm{ 94.37 \pm 0.17 } $ &  $ 90.13 \pm 4.54 $ &  $ 90.28 \pm 5.09 $ \\
              pima &                  $ 76.71 \pm 0.68 $ &  $ \bm{ 76.84 \pm 0.65 } $ &  $ 65.49 \pm 4.08 $ &  $ 69.02 \pm 2.52 $ \\
          satimage &                  $ 82.78 \pm 0.21 $ &  $ \bm{ 82.99 \pm 0.21 } $ &  $ 78.19 \pm 2.00 $ &  $ 77.18 \pm 2.69 $ \\
           segment &                  $ 90.42 \pm 0.27 $ &  $ \bm{ 90.60 \pm 0.37 } $ &  $ 79.25 \pm 3.19 $ &  $ 83.42 \pm 2.96 $ \\
           shuttle &                  $ 93.87 \pm 0.33 $ &  $ \bm{ 93.95 \pm 0.31 } $ &  $ 88.44 \pm 3.55 $ &  $ 87.04 \pm 3.27 $ \\
           vehicle &                  $ 70.63 \pm 0.75 $ &  $ \bm{ 73.57 \pm 1.02 } $ &  $ 54.13 \pm 4.66 $ &  $ 58.97 \pm 4.34 $ \\
         vertebral &                  $ 80.48 \pm 1.11 $ &  $ \bm{ 82.15 \pm 1.00 } $ &  $ 71.79 \pm 1.74 $ &  $ 76.45 \pm 2.31 $ \\
             vowel &                  $ 51.26 \pm 1.78 $ &  $ \bm{ 53.87 \pm 1.17 } $ &  $ 29.61 \pm 1.43 $ &  $ 33.71 \pm 1.69 $ \\
              wdbc &                  $ 97.30 \pm 0.35 $ &  $ \bm{ 97.38 \pm 0.35 } $ &  $ 94.38 \pm 1.73 $ &  $ 93.10 \pm 2.93 $ \\
             yeast &  $ \bm{ 57.37 \pm 0.57 } $ &                  $ 56.18 \pm 1.11 $ &  $ 33.26 \pm 1.94 $ &  $ 43.65 \pm 2.54 $ \\
              \textbf{Mean} &                                  78.02 &            $ \bm{78.60} $ &                  68.39 &                  69.94 \\
\textbf{p-value} & - & 0.0602 & $3.8147 \times 10^{-6}$ & $3.8147 \times 10^{-6}$ \\
\bottomrule
\end{tabular}

}
\end{table}


The p-value indicates that, when there is no noise added to the training instances, the proposed method is statistically equivalent to Bagging in the noise free scenario. These results fall in line with our previous expectations. The proposed method was specifically designed with noisy data sets in mind, or data sets with a large fraction of outliers. As such, it is expected that it will show its best results under these conditions, which the first experiment does not satisfy.

These results indicate that the pools generated by our algorithm achieve results similar to the ones obtained by the Bagging algorithm. This is an indirect but important piece of evidence that points towards the correctness of one of our initial assumptions: that the kDN hardness measure is effective in distinguishing noisy data from noise-free data. This is because, should the kDN measure consider noise-free data points as hard, we would have bootstrapped sets were these data points would be underrepresented, something that would not happen under the original Bagging algorithm. In that case, we would expect to see the proposed method achieve much lower accuracy than that of the original Bagging algorithm.

Furthermore, in this noise-free context, we can also note that both Bagging and our proposed method offer performance gains over the Perceptron algorithm. In contrast, the Random Subspace algorithm shows a lower value for the average accuracy than the Perceptron algorithm in all but six of the data sets, and also larger values for the standard deviation, which in turn indicates high variability of the performance. This may caused by too drastic a reduction of the feature set, considering our data sets have relatively low-dimensional feature sets.

\subsubsection{Noisy Scenarios}

Having established a baseline of performance, we now analyze the noisy scenarios. Tables \ref{tbl:acc10} through \ref{tbl:acc40} present the results for these cases.

\begin{table}[]
  \caption{Mean and standard deviation of the mean accuracy over the folds for each classifier, at a noise level of 10\%. The best values (highest mean) are shown in bold. The last line of the table shows the p-value for each pairwise Wilcoxon test.}
  \label{tbl:acc10}
\resizebox{\columnwidth}{!}{%
\begin{tabular}{lllll}
\toprule
              Dataset &                       \textbf{\textbf{Bagging}-IH} &                                \textbf{Bagging} &         \textbf{Rand. Subsp.} &                  \textbf{Perceptron OvA} \\
\midrule
 blood transfusion &  $ \bm{ 78.36 \pm 0.57 } $ &                  $ 78.18 \pm 0.47 $ &   $ 64.13 \pm 9.40 $ &   $ 64.36 \pm 9.94 $ \\
               ctg &  $ \bm{ 87.92 \pm 0.32 } $ &                  $ 87.32 \pm 0.43 $ &  $ 75.19 \pm 10.66 $ &   $ 76.16 \pm 8.83 $ \\
             ecoli &  $ \bm{ 76.08 \pm 1.31 } $ &                  $ 75.91 \pm 1.11 $ &   $ 53.12 \pm 7.27 $ &   $ 56.85 \pm 6.81 $ \\
             glass &  $ \bm{ 56.42 \pm 2.81 } $ &                  $ 54.99 \pm 2.02 $ &   $ 41.17 \pm 4.79 $ &   $ 43.52 \pm 4.75 $ \\
          haberman &  $ \bm{ 74.14 \pm 0.84 } $ &                  $ 74.02 \pm 1.46 $ &   $ 63.95 \pm 9.82 $ &   $ 62.58 \pm 8.89 $ \\
              ilpd &  $ \bm{ 70.65 \pm 1.07 } $ &                  $ 70.34 \pm 1.01 $ &   $ 61.87 \pm 8.15 $ &   $ 59.81 \pm 7.31 $ \\
        ionosphere &                  $ 83.86 \pm 1.30 $ &  $ \bm{ 85.50 \pm 1.11 } $ &   $ 71.82 \pm 8.32 $ &   $ 70.01 \pm 9.04 $ \\
             liver &  $ \bm{ 64.97 \pm 1.60 } $ &                  $ 64.55 \pm 1.90 $ &   $ 58.19 \pm 1.08 $ &   $ 60.10 \pm 2.62 $ \\
       new\_thyroid &                  $ 90.91 \pm 1.12 $ &  $ \bm{ 90.93 \pm 1.66 } $ &   $ 83.91 \pm 7.24 $ &   $ 86.12 \pm 5.44 $ \\
       page\_blocks &  $ \bm{ 93.96 \pm 0.20 } $ &                  $ 93.55 \pm 0.18 $ &  $ 81.92 \pm 10.05 $ &  $ 82.03 \pm 10.22 $ \\
              pima &  $ \bm{ 76.53 \pm 0.69 } $ &                  $ 75.98 \pm 0.67 $ &   $ 61.80 \pm 5.72 $ &   $ 64.94 \pm 4.73 $ \\
          satimage &  $ \bm{ 82.61 \pm 0.17 } $ &                  $ 80.85 \pm 0.30 $ &   $ 66.82 \pm 6.95 $ &   $ 64.64 \pm 7.43 $ \\
           segment &  $ \bm{ 90.34 \pm 0.30 } $ &                  $ 89.80 \pm 0.52 $ &   $ 68.34 \pm 5.40 $ &   $ 72.29 \pm 7.23 $ \\
           shuttle &  $ \bm{ 93.53 \pm 0.28 } $ &                  $ 90.58 \pm 0.41 $ &  $ 77.80 \pm 11.46 $ &  $ 77.02 \pm 10.52 $ \\
           vehicle &                  $ 70.65 \pm 1.04 $ &  $ \bm{ 73.06 \pm 1.27 } $ &   $ 48.76 \pm 6.10 $ &   $ 54.78 \pm 5.87 $ \\
         vertebral &  $ \bm{ 80.29 \pm 1.39 } $ &                  $ 80.03 \pm 1.68 $ &   $ 67.40 \pm 4.86 $ &   $ 72.74 \pm 5.53 $ \\
             vowel &  $ \bm{ 50.19 \pm 1.03 } $ &                  $ 49.82 \pm 1.48 $ &   $ 25.55 \pm 2.53 $ &   $ 29.33 \pm 2.82 $ \\
              wdbc &  $ \bm{ 97.00 \pm 0.59 } $ &                  $ 95.98 \pm 0.79 $ &   $ 86.08 \pm 7.97 $ &   $ 83.11 \pm 6.81 $ \\
             yeast &  $ \bm{ 57.05 \pm 0.59 } $ &                  $ 55.03 \pm 0.99 $ &   $ 29.78 \pm 4.10 $ &   $ 38.92 \pm 3.53 $ \\
              \textbf{Mean} &            $ \bm{77.66} $ &                                  77.18 &                   62.51 &                   64.17 \\
\textbf{p-value} & - & 0.0124 & $3.8147 \times 10^{-6}$ & $3.8147 \times 10^{-6}$ \\
\bottomrule
\end{tabular}

}
\end{table}

\begin{table}[]
  \caption{Mean and standard deviation of the mean accuracy over the folds for each classifier, at a noise level of 20\%. The best values (highest mean) are shown in bold. The last line of the table shows the p-value for each pairwise Wilcoxon test.}
  \label{tbl:acc20}
\resizebox{\columnwidth}{!}{%
\begin{tabular}{lllll}
\toprule
              Dataset &                       \textbf{\textbf{Bagging}-IH} &                                \textbf{Bagging} &         \textbf{Rand. Subsp.} &                  \textbf{Perceptron OvA} \\
\midrule
 blood transfusion &  $ \bm{ 77.98 \pm 0.73 } $ &                  $ 77.47 \pm 1.10 $ &  $ 59.48 \pm 12.18 $ &  $ 59.55 \pm 12.72 $ \\
               ctg &  $ \bm{ 87.55 \pm 0.43 } $ &                  $ 86.18 \pm 0.53 $ &  $ 63.33 \pm 14.25 $ &  $ 64.82 \pm 15.65 $ \\
             ecoli &  $ \bm{ 76.10 \pm 1.54 } $ &                  $ 74.50 \pm 1.35 $ &   $ 45.44 \pm 8.94 $ &   $ 49.23 \pm 7.81 $ \\
             glass &  $ \bm{ 56.92 \pm 2.90 } $ &                  $ 54.30 \pm 2.31 $ &   $ 37.28 \pm 5.93 $ &   $ 39.38 \pm 5.97 $ \\
          haberman &  $ \bm{ 74.23 \pm 1.35 } $ &                  $ 73.27 \pm 1.86 $ &   $ 63.09 \pm 8.98 $ &   $ 62.93 \pm 8.88 $ \\
              ilpd &  $ \bm{ 69.83 \pm 1.44 } $ &                  $ 68.68 \pm 1.87 $ &   $ 60.98 \pm 8.27 $ &   $ 59.88 \pm 8.09 $ \\
        ionosphere &                  $ 82.52 \pm 1.39 $ &  $ \bm{ 83.46 \pm 1.53 } $ &   $ 63.48 \pm 9.96 $ &   $ 64.65 \pm 6.77 $ \\
             liver &                  $ 61.67 \pm 2.30 $ &  $ \bm{ 61.68 \pm 1.94 } $ &   $ 57.17 \pm 3.47 $ &   $ 58.43 \pm 2.85 $ \\
       new\_thyroid &  $ \bm{ 89.86 \pm 1.77 } $ &                  $ 89.02 \pm 2.17 $ &  $ 77.44 \pm 16.30 $ &  $ 81.02 \pm 12.82 $ \\
       page\_blocks &  $ \bm{ 93.76 \pm 0.22 } $ &                  $ 93.09 \pm 0.27 $ &  $ 68.84 \pm 20.76 $ &  $ 73.00 \pm 18.94 $ \\
              pima &  $ \bm{ 75.38 \pm 1.25 } $ &                  $ 74.45 \pm 1.18 $ &   $ 58.72 \pm 4.49 $ &   $ 61.78 \pm 4.02 $ \\
          satimage &  $ \bm{ 82.25 \pm 0.18 } $ &                  $ 79.63 \pm 0.65 $ &   $ 59.11 \pm 7.80 $ &   $ 56.04 \pm 5.87 $ \\
           segment &  $ \bm{ 90.34 \pm 0.43 } $ &                  $ 87.69 \pm 0.81 $ &   $ 58.31 \pm 5.47 $ &   $ 61.01 \pm 6.14 $ \\
           shuttle &  $ \bm{ 92.36 \pm 0.38 } $ &                  $ 89.44 \pm 0.61 $ &  $ 69.03 \pm 13.36 $ &  $ 68.74 \pm 12.65 $ \\
           vehicle &                  $ 70.51 \pm 1.15 $ &  $ \bm{ 70.93 \pm 1.44 } $ &   $ 42.45 \pm 5.71 $ &   $ 48.39 \pm 6.25 $ \\
         vertebral &  $ \bm{ 78.77 \pm 2.29 } $ &                  $ 78.71 \pm 2.79 $ &  $ 62.45 \pm 11.30 $ &  $ 65.87 \pm 11.46 $ \\
             vowel &  $ \bm{ 49.28 \pm 1.46 } $ &                  $ 46.27 \pm 1.18 $ &   $ 21.94 \pm 3.58 $ &   $ 25.68 \pm 2.69 $ \\
              wdbc &  $ \bm{ 95.86 \pm 0.88 } $ &                  $ 93.70 \pm 0.84 $ &  $ 78.40 \pm 11.42 $ &   $ 76.78 \pm 8.40 $ \\
             yeast &  $ \bm{ 57.23 \pm 0.65 } $ &                  $ 53.99 \pm 1.32 $ &   $ 26.47 \pm 4.65 $ &   $ 35.25 \pm 6.21 $ \\
              \textbf{Mean} &            $ \bm{76.97} $ &                                  75.60 &                   56.50 &                   58.55 \\
\textbf{p-value} & - & 0.0003 & $3.8147 \times 10^{-6}$ & $3.8147 \times 10^{-6}$ \\
\bottomrule
\end{tabular}

}
\end{table}

\begin{table}[]
  \caption{Mean and standard deviation of the mean accuracy over the folds for each classifier, at a noise level of 30\%. The best values (highest mean) are shown in bold. The last line of the table shows the p-value for each pairwise Wilcoxon test.}
  \label{tbl:acc30}
\resizebox{\columnwidth}{!}{%
\begin{tabular}{lllll}
\toprule
              Dataset &                       \textbf{\textbf{Bagging}-IH} &                \textbf{Bagging} &         \textbf{Rand. Subsp.} &                  \textbf{Perceptron OvA} \\
\midrule
 blood transfusion &  $ \bm{ 76.43 \pm 1.48 } $ &  $ 74.70 \pm 2.41 $ &  $ 57.79 \pm 13.41 $ &  $ 57.57 \pm 12.53 $ \\
               ctg &  $ \bm{ 86.56 \pm 0.45 } $ &  $ 85.15 \pm 0.59 $ &  $ 53.32 \pm 15.05 $ &  $ 56.16 \pm 14.81 $ \\
             ecoli &  $ \bm{ 74.99 \pm 1.58 } $ &  $ 72.53 \pm 1.86 $ &  $ 39.91 \pm 10.51 $ &   $ 44.65 \pm 8.92 $ \\
             glass &  $ \bm{ 53.88 \pm 2.82 } $ &  $ 52.29 \pm 2.63 $ &   $ 32.33 \pm 7.74 $ &   $ 33.93 \pm 7.06 $ \\
          haberman &  $ \bm{ 72.86 \pm 1.82 } $ &  $ 70.69 \pm 3.53 $ &  $ 56.03 \pm 12.66 $ &  $ 56.97 \pm 10.72 $ \\
              ilpd &  $ \bm{ 68.56 \pm 1.82 } $ &  $ 66.45 \pm 2.52 $ &   $ 57.72 \pm 9.69 $ &   $ 56.84 \pm 8.08 $ \\
        ionosphere &  $ \bm{ 81.04 \pm 1.90 } $ &  $ 79.15 \pm 3.27 $ &   $ 58.64 \pm 8.48 $ &   $ 58.79 \pm 8.81 $ \\
             liver &  $ \bm{ 58.58 \pm 2.49 } $ &  $ 57.88 \pm 3.55 $ &   $ 55.70 \pm 5.43 $ &   $ 55.96 \pm 4.94 $ \\
       new\_thyroid &  $ \bm{ 88.70 \pm 2.47 } $ &  $ 87.42 \pm 2.73 $ &  $ 64.88 \pm 25.69 $ &  $ 66.91 \pm 25.77 $ \\
       page\_blocks &  $ \bm{ 93.55 \pm 0.18 } $ &  $ 92.81 \pm 0.28 $ &  $ 61.95 \pm 19.32 $ &  $ 63.76 \pm 17.50 $ \\
              pima &  $ \bm{ 73.36 \pm 1.08 } $ &  $ 70.93 \pm 2.59 $ &   $ 54.54 \pm 5.57 $ &   $ 58.01 \pm 5.90 $ \\
          satimage &  $ \bm{ 81.66 \pm 0.32 } $ &  $ 78.23 \pm 0.60 $ &   $ 50.89 \pm 8.27 $ &   $ 50.19 \pm 7.33 $ \\
           segment &  $ \bm{ 90.00 \pm 0.29 } $ &  $ 85.70 \pm 0.78 $ &   $ 50.18 \pm 6.49 $ &   $ 52.52 \pm 7.79 $ \\
           shuttle &  $ \bm{ 91.44 \pm 0.43 } $ &  $ 87.98 \pm 0.54 $ &  $ 59.00 \pm 15.03 $ &  $ 58.49 \pm 15.22 $ \\
           vehicle &  $ \bm{ 69.87 \pm 1.11 } $ &  $ 68.68 \pm 1.78 $ &   $ 37.12 \pm 6.28 $ &   $ 45.96 \pm 6.27 $ \\
         vertebral &  $ \bm{ 74.95 \pm 2.59 } $ &  $ 74.13 \pm 2.71 $ &  $ 62.50 \pm 11.03 $ &  $ 63.87 \pm 10.03 $ \\
             vowel &  $ \bm{ 47.47 \pm 1.54 } $ &  $ 42.42 \pm 1.86 $ &   $ 18.17 \pm 3.14 $ &   $ 22.81 \pm 3.69 $ \\
              wdbc &  $ \bm{ 92.25 \pm 1.96 } $ &  $ 88.44 \pm 2.31 $ &  $ 70.98 \pm 11.10 $ &   $ 66.97 \pm 8.66 $ \\
             yeast &  $ \bm{ 56.45 \pm 0.70 } $ &  $ 52.82 \pm 1.35 $ &   $ 22.83 \pm 5.94 $ &   $ 29.37 \pm 6.91 $ \\
              \textbf{Mean} &            $ \bm{75.40} $ &                  73.07 &                   50.76 &                   52.62 \\
\textbf{p-value} & - & $3.8147 \times 10^{-6}$ & $3.8147 \times 10^{-6}$ & $3.8147 \times 10^{-6}$ \\
\bottomrule
\end{tabular}

}
\end{table}

\begin{table}[]
  \caption{Mean and standard deviation of the mean accuracy over the folds for each classifier, at a noise level of 40\%. The best values (highest mean) are shown in bold. The last line of the table shows the p-value for each pairwise Wilcoxon test.}
  \label{tbl:acc40}
\resizebox{\columnwidth}{!}{%
\begin{tabular}{lllll}
\toprule
              Dataset &                       \textbf{\textbf{Bagging}-IH} &                \textbf{Bagging} &         \textbf{Rand. Subsp.} &                  \textbf{Perceptron OvA} \\
\midrule
 blood transfusion &  $ \bm{ 70.29 \pm 7.14 } $ &  $ 66.22 \pm 8.21 $ &  $ 53.93 \pm 12.92 $ &  $ 55.37 \pm 11.66 $ \\
               ctg &  $ \bm{ 85.14 \pm 0.78 } $ &  $ 82.48 \pm 1.51 $ &  $ 43.92 \pm 19.16 $ &  $ 50.43 \pm 16.28 $ \\
             ecoli &  $ \bm{ 73.68 \pm 2.36 } $ &  $ 69.82 \pm 2.32 $ &   $ 33.00 \pm 8.88 $ &   $ 37.38 \pm 6.50 $ \\
             glass &  $ \bm{ 51.91 \pm 3.50 } $ &  $ 49.52 \pm 2.69 $ &   $ 28.58 \pm 8.13 $ &   $ 29.65 \pm 7.56 $ \\
          haberman &  $ \bm{ 66.18 \pm 5.37 } $ &  $ 61.95 \pm 7.03 $ &  $ 53.98 \pm 14.20 $ &  $ 54.18 \pm 12.89 $ \\
              ilpd &  $ \bm{ 61.21 \pm 3.68 } $ &  $ 60.38 \pm 3.18 $ &   $ 56.55 \pm 9.82 $ &   $ 55.24 \pm 9.58 $ \\
        ionosphere &  $ \bm{ 71.43 \pm 6.10 } $ &  $ 65.85 \pm 7.75 $ &   $ 51.82 \pm 6.59 $ &   $ 52.44 \pm 6.98 $ \\
             liver &  $ \bm{ 54.33 \pm 2.72 } $ &  $ 54.09 \pm 3.38 $ &   $ 52.25 \pm 7.41 $ &   $ 53.39 \pm 5.33 $ \\
       new\_thyroid &  $ \bm{ 86.44 \pm 3.66 } $ &  $ 82.58 \pm 5.34 $ &  $ 53.53 \pm 28.03 $ &  $ 56.28 \pm 27.25 $ \\
       page\_blocks &  $ \bm{ 93.19 \pm 0.28 } $ &  $ 92.41 \pm 0.27 $ &  $ 48.01 \pm 22.47 $ &  $ 50.33 \pm 22.31 $ \\
              pima &  $ \bm{ 66.41 \pm 3.78 } $ &  $ 63.62 \pm 3.95 $ &   $ 52.40 \pm 5.36 $ &   $ 53.86 \pm 5.48 $ \\
          satimage &  $ \bm{ 80.63 \pm 0.37 } $ &  $ 75.92 \pm 1.15 $ &   $ 42.94 \pm 8.64 $ &   $ 41.10 \pm 8.32 $ \\
           segment &  $ \bm{ 89.18 \pm 0.46 } $ &  $ 83.35 \pm 0.97 $ &   $ 40.76 \pm 7.16 $ &   $ 45.08 \pm 7.69 $ \\
           shuttle &  $ \bm{ 90.71 \pm 0.54 } $ &  $ 86.95 \pm 0.68 $ &  $ 46.73 \pm 16.94 $ &  $ 48.33 \pm 18.41 $ \\
           vehicle &  $ \bm{ 67.90 \pm 1.80 } $ &  $ 64.22 \pm 2.41 $ &   $ 32.24 \pm 4.08 $ &   $ 37.99 \pm 4.98 $ \\
         vertebral &  $ \bm{ 67.65 \pm 3.96 } $ &  $ 64.23 \pm 4.74 $ &  $ 59.84 \pm 14.19 $ &  $ 60.60 \pm 11.84 $ \\
             vowel &  $ \bm{ 44.69 \pm 1.87 } $ &  $ 38.18 \pm 2.18 $ &   $ 15.66 \pm 3.16 $ &   $ 19.60 \pm 3.09 $ \\
              wdbc &  $ \bm{ 80.53 \pm 5.42 } $ &  $ 74.19 \pm 5.15 $ &  $ 60.66 \pm 10.26 $ &   $ 59.43 \pm 8.04 $ \\
             yeast &  $ \bm{ 55.93 \pm 0.83 } $ &  $ 50.42 \pm 1.84 $ &   $ 20.32 \pm 7.01 $ &   $ 27.18 \pm 7.49 $ \\
              \textbf{Mean} &            $ \bm{71.44} $ &                  67.70 &                   44.59 &                   46.73 \\
\textbf{p-value} & - & $3.8147 \times 10^{-6}$ & $3.8147 \times 10^{-6}$ & $3.8147 \times 10^{-6}$ \\
\bottomrule
\end{tabular}

}
\end{table}

When analyzing the results for noisy scenarios, the observed behavior differs from that of the noise-free scenario, and the proposed method tends to have the highest mean accuracy. Our method significantly outperforms the other ensemble algorithms tested, and in particular it significantly outperforms the original Bagging algorithm for all noise levels, indicating that the modifications we proposed for dealing with noisy data resulted in a performance gain.

\begin{figure}[]
\centering

  \includegraphics[width=\columnwidth,keepaspectratio]{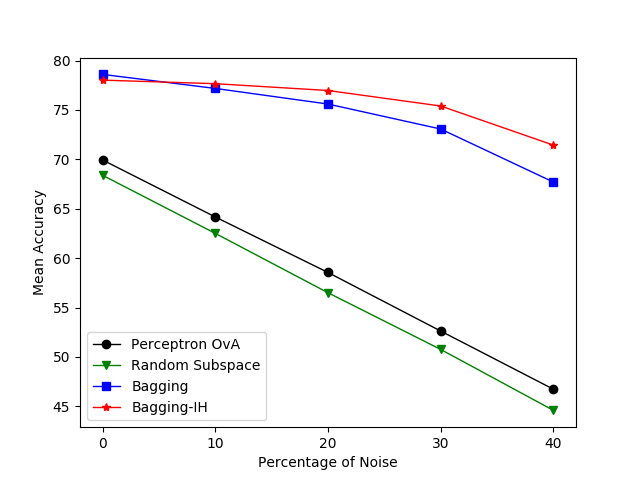}
  \caption{The noise level versus the mean accuracy of each method.}
  \label{fig:accvnoise}

\end{figure}

One can also observe the accuracy of all methods decrease, as the noise level increases, as shown in Figure \ref{fig:accvnoise}. This was to be expected, since the quality of the training data becomes poorer and poorer with each increase in the noise level.

Table \ref{tbl:wilcoxonsummary} summarizes the results of the pairwise Wilcoxon tests. A tilde ($\sim$) indicates that there was not sufficient evidence to reject the null hypothesis, and that our classifier is statistically equivalent to the compared classifier. A plus sign indicate that the null hypothesis was rejected, and our classifier is significantly better than the compared classifier. The results in Table \ref{tbl:wilcoxonsummary} suggest that our method is indeed better suited to dealing with noisy data than the other algorithms it was compared against, given that we achieve significantly higher accuracy than the other methods in noisy scenarios.

\begin{table}[]
\centering
\caption{Summary of results for the paired Wilcoxon tests. A tilde indicate that our method is statistically equivalent to the compared classifier, whereas a plus sign indicates that our method is significantly better.}
\label{tbl:wilcoxonsummary}

\resizebox{\columnwidth}{!}
{
\begin{tabular}{@{}llll@{}}
\toprule
\textbf{Noise Level} & \textbf{Bagging} & \textbf{Random Subspace} & \textbf{Perceptron OvA} \\ \midrule
0\%                  & $\sim$           & +                        & +                       \\
10\%                 & +                & +                        & +                       \\
20\%                 & +                & +                        & +                       \\
30\%                 & +                & +                        & +                       \\
40\%                 & +                & +                        & +                       \\ \bottomrule
\end{tabular}
}
\end{table}

\subsubsection{Frequency of Noisy Instances}

Beyond just calculating the accuracy for each noise level, we also calculated for every fold the average frequency with which noisy instances were added to the bootstrapped sets. That is, for every fold, we measured the relative frequency of noisy instances in each of the $m$ bootstrapped sets, and averaged those relative frequencies to obtain a single value, denoted $freq_{noisy}$. We then observe the distribution of values for $freq_{noisy}$, and compare that to the expected value for the original Bagging algorithm. Note that, for the Bagging algorithm, at a noise level of $pr_{change}$, we expect a fraction equal to $pr_{change}$ of the instances in each bootstrapped set will be noisy instances, on average. This is due to the fact that each instance is equally likely to be selected.

\begin{figure}[h]
  \centering
  \includegraphics[width=\columnwidth,keepaspectratio]{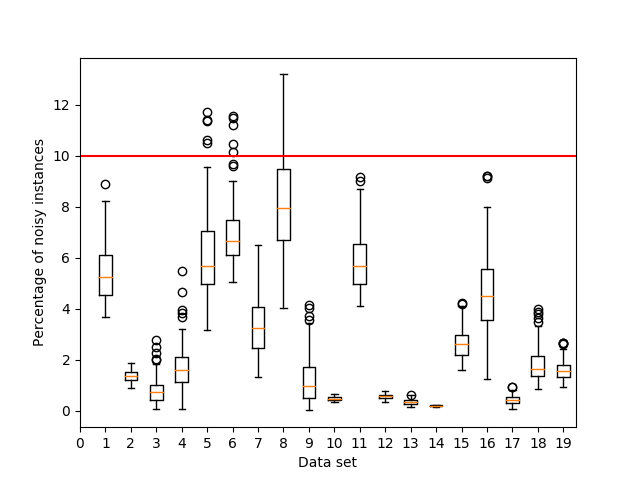}
  \caption{Boxplot of the frequency of noisy instances added to the bootstrapped sets, at a noise level of 10\%. The horizontal line shows the expected frequency of noisy instances selected by the original Bagging algorithm.}
  \label{fig:boxplot01}

\end{figure}

\begin{figure}[h]
\centering

  \includegraphics[width=\columnwidth,keepaspectratio]{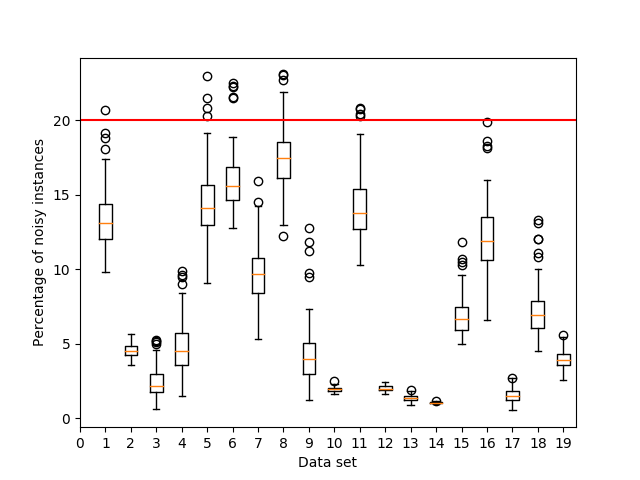}
  \caption{Boxplot of the frequency of noisy instances added to the bootstrapped sets, at a noise level of 20\%. The horizontal line shows the expected frequency of noisy instances selected by the original Bagging algorithm.}
  \label{fig:boxplot02}

\end{figure}

\begin{figure}[h]
\centering

  \includegraphics[width=\columnwidth,keepaspectratio]{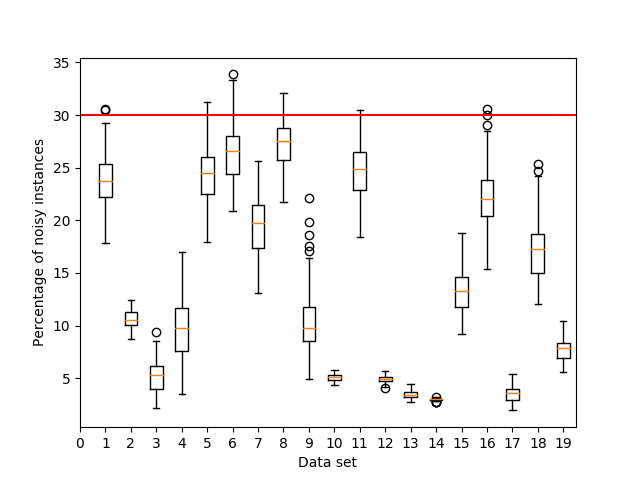}
  \caption{Boxplot of the frequency of noisy instances added to the bootstrapped sets, at a noise level of 30\%. The horizontal line shows the expected frequency of noisy instances selected by the original Bagging algorithm.}
  \label{fig:boxplot03}

\end{figure}

\begin{figure}[h]
\centering

  \includegraphics[width=\columnwidth,keepaspectratio]{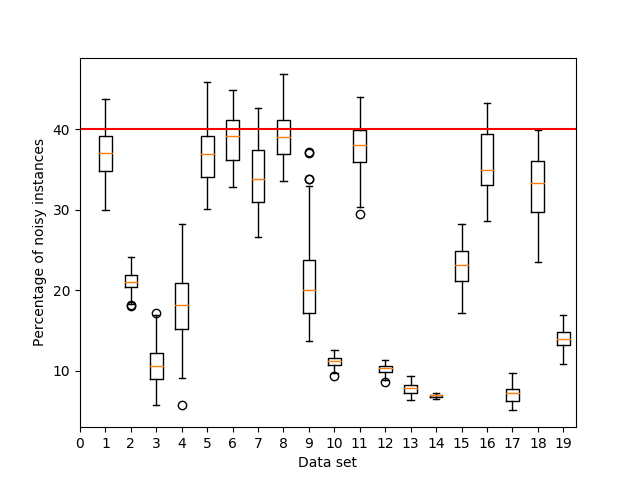}
  \caption{Boxplot of the frequency of noisy instances added to the bootstrapped sets, at a noise level of 40\%. The horizontal line shows the expected frequency of noisy instances selected by the original Bagging algorithm.}
  \label{fig:boxplot04}

\end{figure}

Figures \ref{fig:boxplot01} through \ref{fig:boxplot04} show the box plot of the values of the frequency of noisy instances selected. Each box plot contains 100 data points, 5 (one per fold) for each of the 20 repetitions of the experiment. The horizontal line in each figure represents the expected percentage of noisy instances selected by the original Bagging algorithm at each noise level. This fraction is simply the noise level, since the Bagging algorithm selects instances by uniform random sampling.

The results show that, for most datasets and most noise levels, nearly all values for the frequency of noisy instances lie below the expected value for the original Bagging algorithm. This is particularly true for the lowest noise levels. Figure \ref{fig:boxplot01} shows that for every data set except for the liver data set, the upper adjacent of the distribution of values always lies below the expected frequency of 10\% for the original Bagging algorithm. The liver dataset (\#8), however, appears to be an outlier in its behavior, since it is the only one where we observe such a distribution of values for every noise level.

Even at the highest noise level of 40\%, in most data sets the frequency of noisy instances picked is still lower than the expected 40\% at least 75\% of the time. The exceptions are the pima (\#11), liver (\#8) and ilpd (\#11) data sets. Even in these cases, the frequency of noisy instances is lower than 40\% in more than 50\% of the total folds.

We do not observe a clear link between the cases in which Bagging outperforms our method in Tables \ref{tbl:acc10} and \ref{tbl:acc20} and anomalously high percentages of noisy instances being selected. In all three datasets in which Bagging achieves higher mean accuracy, we do not observe noisy instances being selected by our method more often than Bagging would have selected noisy instances. This indicates that perhaps, for these three datasets, our method may have been to eager to filter instances, but further experiments are needed to verify this.

These results offer support to our initial hypothesis that the kDN hardness measure is capable of discriminating between noisy and noise-free instances. Moreover, it also validates our method, in that it shows that our method is capable of choosing mostly noise free instances to be part of the bootstrapped sets, which should lead to better performance.

\FloatBarrier

\section{Conclusions and Future Work}\label{sec:conclusion}

In this work, we proposed a new method which combines data complexity measures and ensemble methods to achieve better classification accuracy on scenarios which involve noisy data. More specifically, our method leveraged the k-Disagreeing Neighbors measure of instance hardness to modify the bootstrapping process of the Bagging algorithm.

The proposed modification to Bagging aimed to avoid adding noisy examples to the training datasets as much as possible, since several works in the classification literature pointed to the adverse effects of noise. Nevertheless, we adopted a filtering approach that was probabilistic in nature, in order to preserve examples that are inherently hard, such as examples on the border of classes. To achieve this, we introduced a procedure for calculating instance selection probabilities during the construction of the bootstrapped training sets based on the hardness of the instance.

We performed experiments on 19 publicly available datasets, comparing our method with the Random Subspace and Bagging algorithms, widely used in the literature. We also compared the ensemble algorithms with a single classifier, in order to obtain a baseline of performance.

Our results indicate that our proposed method performs at least as well as Bagging in every scenario tested. Furthermore on experiments with noise levels of 10\% or greater, our algorithm is the best performing method, and there is a significant difference between our method and the Bagging algorithm. These results suggest that our algorithm is better suited than Bagging to dealing with high label noise levels.

The analysis of the distribution of the frequency with which noisy instances were selected offers evidence to support our initial hypothesis that our method would be less prone to selecting noisy instance.

Future work may include an investigation into the use of instance hardness measures other than the k-Disagreeing Neighbors measure. More broadly, it would be interesting to investigate the use of instance hardness measures not based on class labels. This is because these measures are better suited to classification problems, precluding the development of techniques focused on regression. Another possibility for investigation is combining ensemble methods other than Bagging with data complexity measures. Finally, while our work was focused on static combination methods, one could also attempt to combine data complexity measures with dynamic selections schemes.


\section*{Acknowledgment}

The authors would like to thank CAPES (Coordenação de Aperfeiçoamento de Pessoal de Nível Superior, in portuguese), CNPq (Conselho Nacional de Desenvolvimento Científico e Tecnológico, in portuguese) and FACEPE (Fundação de Amparo à Ciência e Tecnologia do Estado de Pernambuco, in portuguese )



\bibliographystyle{IEEEtran}
\bibliography{bibliography}
%
%

\end{document}